\documentclass[10pt,twocolumn,letterpaper]{article}

\usepackage{iccv}
\usepackage{times}
\usepackage{epsfig}
\usepackage{graphicx}
\usepackage{amsmath}
\usepackage{amssymb}
\usepackage{pifont}
\usepackage{comment}

\usepackage{algorithm}
\usepackage{algorithmic}

\usepackage[breaklinks=true,bookmarks=false]{hyperref}

\iccvfinalcopy 

\def\iccvPaperID{5057} 

\ificcvfinal\fi

\begin{document}

\title{Improving Online Lane Graph Extraction by Object-Lane Clustering}

\author{ Yigit Baran Can\textsuperscript{1}\space\space\space\space Alexander Liniger\textsuperscript{1}\space\space\space\space Danda Pani Paudel\textsuperscript{1,3}\space\space\space\space Luc Van Gool\textsuperscript{1,2,3}\\
\textsuperscript{1}Computer Vision Lab, ETH Zurich\space\space\space\space \textsuperscript{2}VISICS, ESAT/PSI, KU Leuven\space\space\space\space \textsuperscript{3}INSAIT, Sofia University \\ {\tt\small $\{$yigit.can, alex.liniger, paudel, vangool$\}$@vision.ee.ethz.ch} }

\maketitle
\ificcvfinal\thispagestyle{empty}\fi

\begin{abstract}
Autonomous driving requires accurate local scene understanding information. To this end, autonomous agents deploy object detection and online BEV lane graph extraction methods as a part of their perception stack. In this work, we propose an architecture and loss formulation to improve the accuracy of local lane graph estimates by using 3D object detection outputs. The proposed method learns to assign the objects to centerlines by considering the centerlines as cluster centers and the objects as data points to be assigned a probability distribution over the cluster centers. This training scheme ensures direct supervision on the relationship between lanes and objects, thus leading to better performance. The proposed method improves lane graph estimation substantially over state-of-the-art methods. The extensive ablations show that our method can achieve significant performance improvements by using the outputs of existing 3D object detection methods. Since our method uses the detection outputs rather than detection method intermediate representations, a single model of our method can use any detection method at test time. The code will be made publicly available.
\end{abstract}

\section{Introduction}

Accurate road scene understanding is essential for autonomous driving. Two of the most important aspects of road scene understanding are lane graph representation and object detection. While the former defines the action environment of the autonomous agent, the latter provides information on the other traffic agents. The resulting representations are crucial for downstream tasks such as predicting the motion of agents~\cite{cui2019multimodal, hong2019rules, rella2021decoder} and planning the ego-motion~\cite{DBLP:conf/rss/BansalKO19, chen2020learning, espinoza22a}. Although object detection has to be carried out online, lane graphs are frequently obtained from offline generated HD-Maps~\cite{jaritz20202d,seif2016autonomous,ma2019exploiting,ravi2018real,casas2021mp3}. However, traffic scenes are highly dynamic and the offline maps have to be complemented by online components \cite{DBLP:conf/icra/LiWWZ22, liu2022vectormapnet,liao2022maptr,wu2023pix2map,xu2022centerlinedet}. Moreover, the geographically limited coverage of HD-Maps severely limits the widespread applicability of autonomous driving.    

Online scene understanding has been studied widely in literature. Some methods focus on Bird's-Eye-View occupancy grid representations \cite{DBLP:conf/cvpr/RoddickC20,can2020understanding,wu2020motionnet,paz2020probabilistic,yang2018hdnet,casas2018intentnet,philion2020lift, zhou2022cross}. These methods combine the onboard sensor information on a BEV grid with each grid location being a semantic vector representing the properties of that location. While BEV semantic segmentation provides crucial information for downstream tasks such as drivable area, walkways and pedestrian crossings; they fail to provide adequate information on the road network. A complementary line of work is extraction of lane boundaries~\cite{DBLP:conf/ivs/NevenBGPG18, DBLP:conf/iccvw/GansbekeBNPG19, homayounfar2019dagmapper}. Building on previous works, HD-Maps aim to provide a more comprehensive understanding of the traffic scene by incorporating lane boundaries as well as other static elements such as pedestrian crossings \cite{DBLP:conf/icra/LiWWZ22, liu2022vectormapnet,liao2022maptr,xu2022centerlinedet}. However, all these lines of work lack the structured representation that the downstream tasks require. Recently, directly estimating the lane graph from onboard sensors was proposed \cite{Can_2021_ICCV, can2022topology}. The lane graph provides instances of centerlines with traffic directions and provides connectivity of these centerlines. 

\begin{figure*}
    \centering
    \includegraphics[width=.9\linewidth]{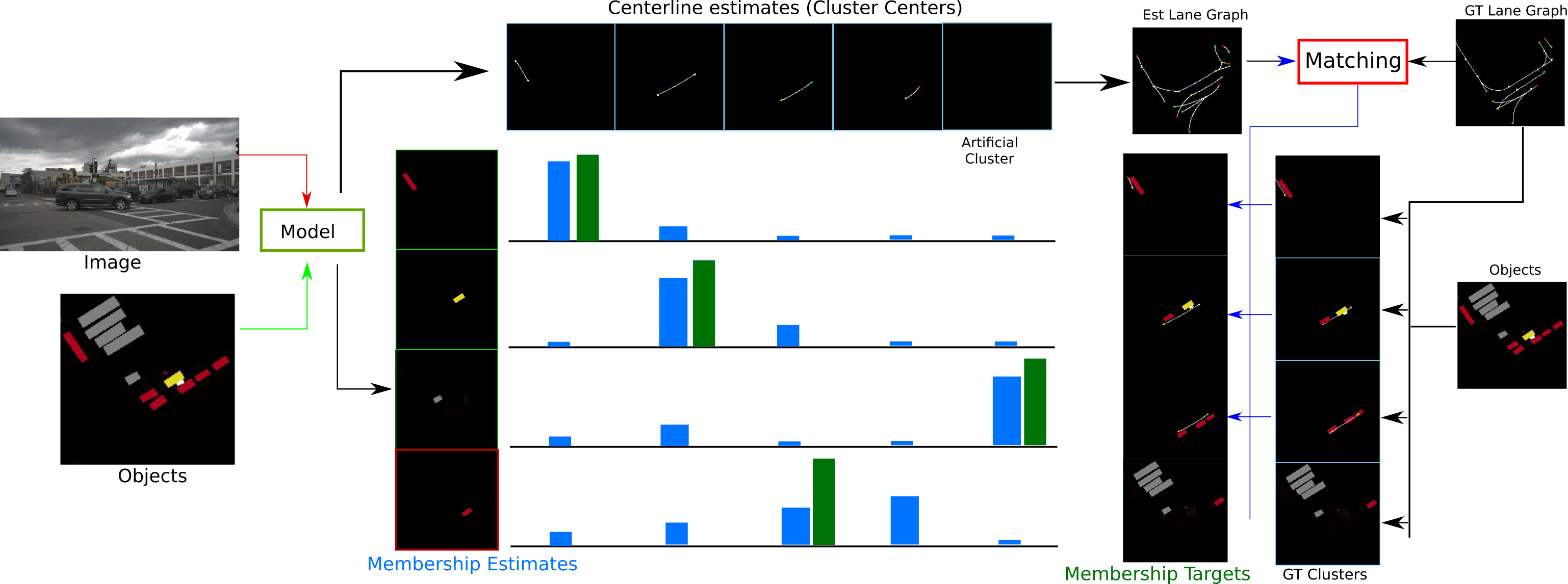}

        \caption{Input image and the object detection estimates are processed jointly to  produce a lane graph with centerlines acting as cluster centers. Each object also outputs a membership probability distribution over the estimated centerlines representing that particular object belonging to estimated centerlines. These membership estimates (blue) are supervised by the targets (green) obtained from true lane graph and object detections. }
    \label{fig:teaser}
 \vspace{-1em}
\end{figure*}

 Some scene understanding methods output object detections as well and it has been shown that the auxiliary task of object detection helps with lane graph extraction performance \cite{Can_2021_ICCV}. In this paper, we focus on a related yet different question: Can the object detection results be used to improve the online lane graph extraction? Given the object detection methods use the same onboard sensors as online scene understanding methods and are an indispensable part of the perception stack, using the outputs of object detection methods creates no overhead. Moreover, using the outputs rather than intermediate representations means lane graph extraction can use different object detection methods in train and test time and does not require re-training as the object detection method changes. To this end, we devise a network architecture that takes the onboard image and a set of 3D bounding boxes as input and outputs the BEV lane graph. Additionally, we propose a novel clustering based formulation where the centerlines act as cluster centers and the object detections are assigned to clusters. The network learns to output membership probability estimates for each object detection over the estimated centerlines, see Fig.~\ref{fig:teaser}. These estimates are supervised by the membership targets produced by true lane graph and object estimates. When trained with the proposed formulation, the proposed method produces substantially improved results over state-of-the-art. To summarize, our major contributions can be listed as follows.
\begin{enumerate}
\setlength{\itemsep}{0pt}
\setlength{\parskip}{0pt}
\item We propose an architecture that uses 3D object detections as inputs and produces lane graphs. 
\item We propose a novel formulation that connects the object detections to centerlines to significantly boost the performance, using an intuitive prior of road scenes.
\item We conduct extensive ablations to validate the design choices as well as the robustness of our method in terms of choice 3D object detection methods.
\item The results obtained by our method are significantly superior to the compared methods in all metrics on benchmark datasets with no runtime sacrifice.
\end{enumerate}

\section{Related Work}

The extraction of lane graphs has been mostly studied under offline paradigms. In offline settings, methods utilize a wide variety of sensor inputs such as aerial images, time aggregated cameras, LiDAR and radar data. Road network extraction from aerial images have improved over time to produce impressive results in terms of lane graph accuracy ~\cite{auclair1999survey, richards1999remote, batra2019improved,sun2019leveraging,ventura2018iterative}. Some methods aggregate information from images and 3D sensors such as LiDAR and radar, over multiple passes through the same region. This allows for the sparse 3D information to be condensed and used in more accurate representations  ~\cite{liang2019convolutional, homayounfar2018hierarchical,liang2018end}. Both aerial images and time aggregated sensor data necessitate offline processing which, in turn, requires an online ego-vehicle localization pipeline to be effectively used in downstream tasks.

A more relevant line of work for this paper focuses on structured representations of the traffic scene, especially road network. Lane boundary detection has been tackled for highways by parameterizing the boundaries through polylines ~\cite{DBLP:conf/cvpr/HomayounfarMLU18}. The polyline control points are estimated by a recursive method. A similar work uses aerial LiDAR scans in an RNN to first estimate initial lane boundary points. The initial points are used in Polygon-RNN ~\cite{DBLP:conf/cvpr/AcunaLKF18} to generate the whole boundary. Both works are restricted to the aerial LiDAR reflectance of highways and fail to generalize to more complicated scenarios with onboard sensors.  

Another relevant literature focuses on task of lane estimation. This task has been studied extensively ~\cite{DBLP:conf/ivs/NevenBGPG18, DBLP:conf/iccvw/GansbekeBNPG19}. While earlier works extract the lanes in the image plane ~\cite{DBLP:conf/iccv/GarnettCPLL19, DBLP:journals/corr/abs-2002-06604}, the more recent works aim to project the image information to BEV and then carry out the processing in BEV ~\cite{DBLP:journals/corr/abs-2011-01535, DBLP:conf/siu/YenIaydinS18,DBLP:conf/ivs/NevenBGPG18}. Recently, a more structured approach has been proposed ~\cite{DBLP:conf/icra/LiWWZ22, liu2022vectormapnet, xu2022centerlinedet, liao2022maptr} to extract lanes as well as pedestrian crossings from camera and LiDAR information. While the representations of these methods provide some structure in their estimates, they still focus on lane boundaries and provide no information on the connectivity of lanes. This representation is not helpful for complicated scenes such as roundabouts and requires further processing or a complementary method to extract lane graphs that are used in downstream tasks.    

Bird's-Eye-View understanding of the traffic scenes has been investigated in the framework of semantic occupancy grids. While some methods focus on using only cameras ~\cite{DBLP:conf/cvpr/RoddickC20,philion2020lift,can2020understanding, zhou2022cross}, some other methods aim to combine information from different sensors \cite{pan2020cross,hendy2020fishing}. While the performance of the BEV segmentation methods has been improving and they provide useful information on the traffic scene, their outputs are mostly used as complementary information alongside lane graphs in downstream tasks. 

The most related line of work is online lane graph extraction from onboard camera images \cite{Can_2021_ICCV, can2022topology}. These methods provide a BEV lane graph in a local region surrounding the ego=vehicle. The output is a directed graph that provides information on centerlines, their connectivities and traffic direction. This representation is crucial for complicated scenes such as roundabouts and crossroads, where the other representations fail to model.

\section{Method}

\subsection{Lane Graph Representation}
\label{lanegraph}

We follow a similar definition to \cite{Can_2021_ICCV} in defining the lane graph as a directed graph. Specifically, let us represent the graph by $X(V,E)$ with $V$ representing the vertices and $E$ representing the edges of the graph. The edges are represented by the incidence matrix $A$ with $A[x,y] = 1$ if the centerlines $x$ and $y$ are connected and 0 otherwise. The centerlines, which represent the vertices of the graph, are modeled by Bezier curves with three control points. 

\subsection{Overall Method}

In this work, we aim to create an architecture that accepts an onboard camera image $I$ and 3D object detections $B$ to output the local lane graph in the Bird's-Eye-View. Furthermore, we want to explicitly couple the objects with the centerlines of the local lane graph to provide a better training framework. The overall method is presented in Fig \ref{fig:teaser}. The method simultaneously outputs the lane graph estimate as well as the probability that a given object belongs to a particular centerline in the estimated lane graph. The method is supervised by the true lane graph and the coupling of the object detections with the centerlines in the true lane graph. This coupling can be considered as a clustering where each centerline acts as a cluster center and the objects are assigned to the centerlines. Our method uses a clustering formulation to optimize the cluster centers (centerlines) such that the given data points (object detections) produce the maximum likelihood. 

The derivation of the proposed formulation results in two terms: a centerline posterior probability and a clustering likelihood term. We convert these terms into suitable loss functions and iteratively optimize model parameters by minimizing the total loss. This provides additional supervision to the model by explicitly modelling the relationship between the objects and the centerlines.

\subsection{Centerlines as Clusters}

Let $I$ represent the input image, $B$ represent the objects and $\mathbf{X}$ represent the lane graph. Here, $I$ and $B$ are given constants while $\mathbf{X}$ is a random variable. Given these definitions, we wish to maximize $P(\mathbf{X}| B, I)$. Then,
\begin{align}
    P(\mathbf{X}| B, I) = \dfrac{P(\mathbf{X}, B| I)}{P(B|I)} = \dfrac{P(B|\mathbf{X}, I)P(\mathbf{X}|I)}{P(B|I)}\,.
    \label{eq:eq1}
\end{align} 
As mentioned, we accept image $I$ and the objects $B$ as given. Therefore,
\begin{align}
    P(\mathbf{X}| B, I) \propto P(B|\mathbf{X}, I)P(\mathbf{X}|I)\,.
    \label{eq:eq2}
\end{align}
Here, we will focus on the term $P(B|\mathbf{X}, I)$ which we will refer to as ``maximum likelihood" or ``ML" term. This term measures the probability of the objects given the image and the lane graph. However, in our problem setup, we do not know the lane graph which is the variable we wish to estimate. Therefore, let us approach this problem as an iterative procedure akin to an expectation-maximization strategy. We define the log likelihood as $L(\mathbf{X}) 	\doteq \text{log}P(B|\mathbf{X}, I)$, and represent the lane graph estimate at iteration $n$ by $X_n$. Now, at iteration $n+1$, the goal is finding an $\mathbf{X}$ such that $P(B|\mathbf{X}, I) > P(B|X_n, I)$, or equivalently, ${L(\mathbf{X}) > L(X_n)}$. In order to reach the desired formulation we start by introducing a latent variable $\mathbf{Z}$ such that,
\begin{equation}
        P(B|\mathbf{X}, I) = \sum_{z} P(B|\mathbf{X}, I, \mathbf{Z})P(\mathbf{Z}|\mathbf{X},I)\,.
    \label{eq:eq3}
\end{equation}
Then, the difference between log likelihoods of the $n$th estimate and the current random variable is given by, 
\begin{equation}
\medmuskip=-2mu
\scriptsize
    L(\mathbf{X})-L(X_n)=\text{log}[\Sigma_{z} P(B|\mathbf{X}, I, \mathbf{Z})P(\mathbf{Z}|\mathbf{X},I)] - \text{log}P(B|X_n, I).
    \label{eq:eq4}
\end{equation}
Applying Jensen's inequality to the log of sum term and after some rearrangements, we arrive at the term,
\begin{align}
    L(\mathbf{X}) - L(X_n) \geq&  \sum_{z}P(\mathbf{Z}|B,X_n,I)\text{log}\!\left( \dfrac{P(B,\mathbf{Z}| \mathbf{X},I)}{P(B,\mathbf{Z}| X_n,I)}\right)  \nonumber\\
    =&\text{log}\dfrac{P(B|\mathbf{X},I)}{P(B|X_n,I)} \nonumber\\
 +& \sum_{z}P(\mathbf{Z}|B,X_n,I) \text{log}\!\left( \dfrac{P(\mathbf{Z}| B, \mathbf{X},I)}{P(\mathbf{Z}| B, X_n,I)}\right) \nonumber\\ 
 =&\text{log}\dfrac{P(B|\mathbf{X},I)}{P(B|X_n,I)} + \phi(\mathbf{X}|X_n).
    \label{eq:eq5}
\end{align}
Combining Eq.~\ref{eq:eq2} with the above result, we obtain,
{\medmuskip=-2mu
\scriptsize
\begin{align}
    L(\mathbf{X}) + \text{log}P(\mathbf{X} | I) & \geq 
 \text{log}\dfrac{P(B|\mathbf{X},I)}{P(B|X_n,I)} + \phi(\mathbf{X}|X_n) +L(X_n) + \text{log}P(\mathbf{X} | I)\nonumber \\=& 
 \text{log}\dfrac{P(\mathbf{X}|B,I)P(X_n|I)}{P(X_n|B,I)P(\mathbf{X}|I)} + \phi(\mathbf{X}|X_n) +L(X_n) + \text{log}P(\mathbf{X} | I)\nonumber  \\ =& 
 \text{log}P(\mathbf{X}|B,I) - \text{log}\dfrac{P(X_n|B,I)}{P(X_n|I)} + \phi(\mathbf{X}|X_n) +L(X_n). 
    \label{eq:eq6}
\end{align}
} We need to maximize the resulting expression over $\mathbf{X}$. Opening  $\phi(\mathbf{X}|X_n)$ and removing the terms independent of  $\mathbf{X}$ gives us the \textbf{total term, TT} in the following expression, 
{\medmuskip=-0mu
\small
\begin{align}
    \mathbf{TT} = \overbrace{\text{log}P(\mathbf{X}|B,I)}^\text{Posterior Term} + \overbrace{\sum_{z}P(\mathbf{Z}|B,X_n,I) \text{log}P(\mathbf{Z}| B, \mathbf{X},I)}^\text{Clustering Term}.
    \label{eq:tempmax}
\end{align}
}

The above \textbf{clustering term} is the negative cross entropy between $P(\mathbf{Z}| B, \mathbf{X},I)$ and $P(\mathbf{Z}|B,X_n,I)$. It increases as $P(\mathbf{Z}| B, \mathbf{X},I)$ approaches $P(\mathbf{Z}|B,X_n,I)$. These two terms are the evaluation of the same function $P(\mathbf{Z}|\mathbf{X},B,I)$ at different points $X_n$ and $\mathbf{X}$. Moreover, this relation holds for any probability function $P(\mathbf{Z}|\mathbf{X},B,I)$.
We aim to formulate a method to update $\mathbf{X}$ that will increase the measure of Eq.~\ref{eq:tempmax}. During the search, we will resort to stochastic gradient descent over the dataset. Therefore, instead of minimizing Eq.~\ref{eq:tempmax} directly, we will update $X_{n+1}= X_n + \partial \sigma/\partial X$. Note that the lane graph estimate $X_n$ is actually a function of input image $I$ and the objects $B$, where the function is a neural network. Let us define the function $X_n=F_{\theta_n}(I, B)$, where $\theta_n$ represent the network parameters at iteration $n$. 
Then, an iteration to increase the total term \textbf{TT} can be written as, 
\begin{align}
      \theta_{n+1}= \theta_n + \partial \mathbf{TT}/\partial \theta.
    \label{eq:probmax}
\end{align}
After considering the fact that we have the lane graph ground-truth labels $\overline{X}$, we can convert the above optimization problem into a supervised learning problem with a standard loss function. The first term of Eq.~\ref{eq:tempmax} is the log of the posterior probability. The loss functions that maximize this term are the ones used in previous works~\cite{Can_2021_ICCV, can2022topology}.


A close inspection shows that the clustering term is similar to the expected value of the log likelihood function in the Expectation-Maximization (EM) paradigm \cite{dempster1977maximum}. Specifically, the set of data points is $B$, the latent variable is $\mathbf{Z}$ and the parameters to maximize are given by $\mathbf{X}$ with the parameter values in the current iteration being $X_n$. The difference is we take expectation of $\text{log}P(\mathbf{Z}| B, \mathbf{X},I)$ instead of joint probability $\text{log}P(\mathbf{Z},B| \mathbf{X},I)$. As mentioned before, similar to EM, we also focus on the `maximum likelihood' term. Inspired by these observations, we propose to interpret $P(\mathbf{Z}| B, \mathbf{X},I)$ as \textbf{membership} probability which is a frequently used interpretation in EM clustering methods such as Gaussian Mixture Models (GMM). Similar to GMM, we will assign each data point $B$ to a cluster where a cluster is given by a centerline in $X$. This interpretation makes both theoretical and practical sense since centerlines define the action space of traffic agents and each road traffic agent (car, truck, etc.) is expected to occupy only one centerline at any given time.

At this point, let us assume that we can calculate a target distribution $P(\overline{Z}|B,\overline{X},X_n, I)$ by using the true lane graph $\overline{X}$. Then, we can improve the clustering term using the cross-entropy loss with the label being $P(\overline{Z}|B,\overline{X},X_n, I)$ and the estimate distribution being $P(\mathbf{Z}| B, \mathbf{X},I)$. Therefore, the new term becomes,
\begin{align}
      \sum_{z}P(\overline{Z}|B,\overline{X},\mathbf{X},I) \text{log}P(\mathbf{Z}| B, \mathbf{X},I).
    \label{eq:newpost}
\end{align}
This term is a negative cross entropy function $H(p,q)$. Therefore, this term naturally leads itself to the cross entropy loss between the estimated $\mathbf{Z}$ and the target probability $\overline{Z}$. In the next subsection, we explain how we obtain a ground truth membership function $\overline{Z}$ such that it can be used to guide the estimated membership.

\subsection{Relation to Expectation-Maximization}

To better understand the clustering term, let us create a new term by adding $P(B|\textbf{X},I)$ to the clustering term,
{\medmuskip=-0mu
\small
\begin{align}
    \sum_{z}P(\mathbf{Z}|B,X_n,I) \text{log}P(\mathbf{Z},B| \mathbf{X},I) = \nonumber \\
    \text{log}P(B| \mathbf{X}, I) + \sum_{z}P(\mathbf{Z}|B,X_n,I) \text{log}P(\mathbf{Z}| B, \mathbf{X},I).
    \label{eq:emtemp}
\end{align}
}

We recognize the term in Eq \ref{eq:emtemp} as the expected value of the log likelihood function in expectation-maximization paradigm. Specifically, the set of data points is $B$, the latent variable is $\mathbf{Z}$ and the parameters to maximize is given by $\mathbf{X}$ with the parameter values in current iteration is $X_n$. Then, let us name the log likelihood term $ \sum_{z}P(\mathbf{Z}|B,X_n,I) \text{log}P(\mathbf{Z},B| \mathbf{X},I) $ as \textbf{EM} term. Then, \textbf{total term} becomes

{\medmuskip=-0mu

\begin{align}
\text{log}P(\mathbf{X}|B,I) -  \text{log}P(B| \mathbf{X}, I) + \mathbf{EM}.
    \label{eq:comb}
\end{align}
}

If we rearrange the terms, we arrive at the following formula:

{\medmuskip=-0mu

\begin{align}
\text{log}P(\mathbf{X}|I) -  \text{log}P(B| I) + \mathbf{EM}.
    \label{eq:comb2}
\end{align}
}

Since $P(B| I)$ is given and does not depend on $\mathbf{X}$, we can remove it from the formula. We see that through our formulation, we actually maximize \textbf{EM} term and $\text{log}P(\mathbf{X}|I)$. Comparing with \ref{eq:eq2}, we see that we directly maximize $P(\mathbf{X}|I)$ and apply expectation maximization on $P(B|\mathbf{X}, I)$. This is evident from the fact that \textbf{EM} term is exactly the log likelihood for data points $B$ and parameters $\mathbf{X}$. 

In practice we feed the object information to the network, thus we actually maximize the \textbf{target distribution TD}:
{\medmuskip=-0mu
\begin{align}
\mathbf{TD} = \text{log}P(\mathbf{X}|I, B) + \mathbf{EM}.
    \label{eq:fin}
\end{align}
}

\textbf{TD} has the traditional posterior term as well as the \textbf{EM} term. As shown in our experiments, EM term provides additional supervision through explicit modelling of the relationship between the objects and the centerlines.


\subsection{Membership Function}
\label{sec:membership}

\begin{figure}
    \centering
    \includegraphics[width=\linewidth]{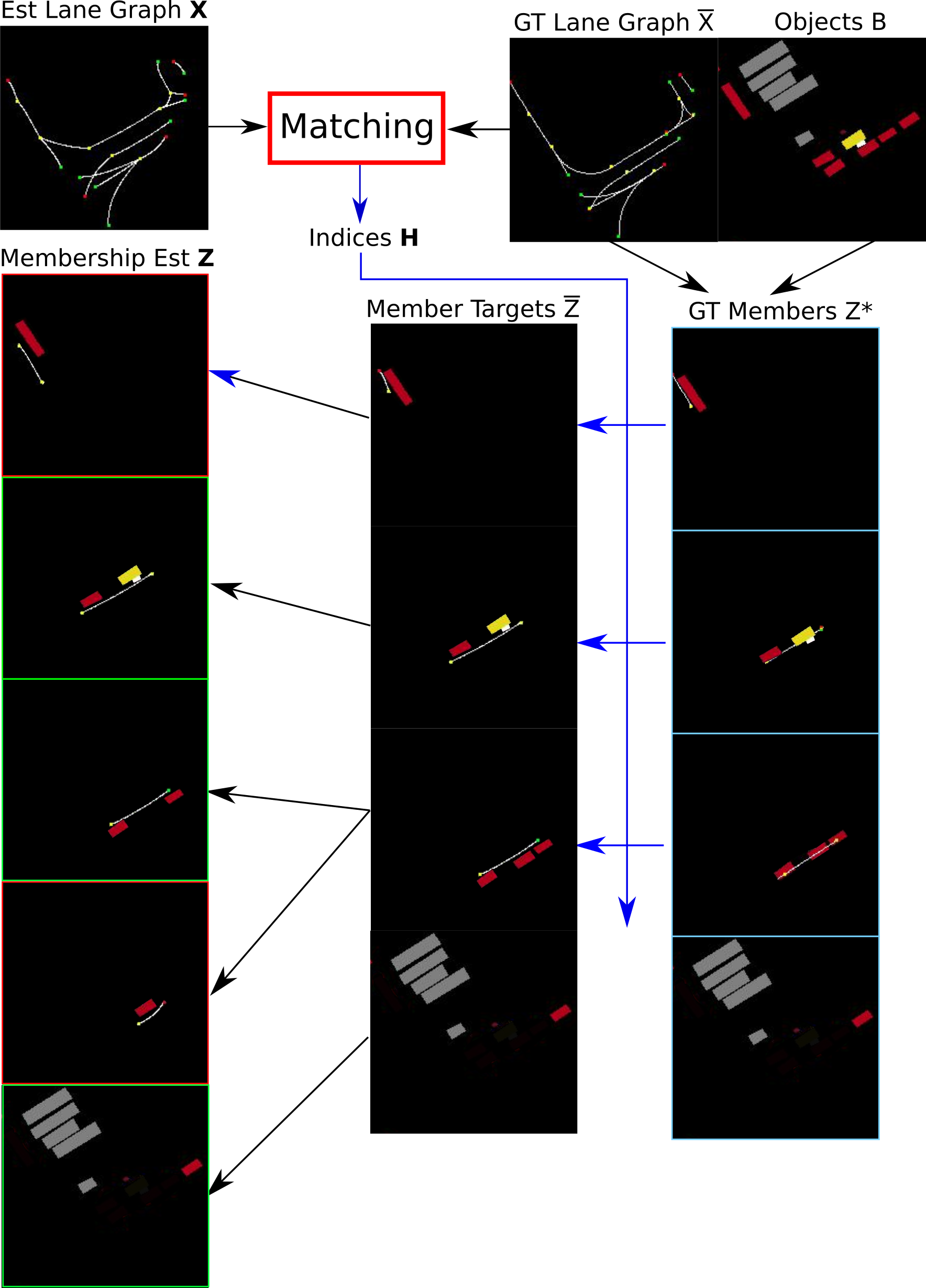}
        \caption{Obtaining true memberships start with matching estimated and GT centerlines, which act as cluster centers. Based on the matching, objects are assigned to the estimated centerlines. Objects that are not matched to any GT centerline are assigned to the outlier set $X_A$. 
        }
    \label{fig:objectmatching}

\end{figure}

\begin{figure*}[h!]
    \centering
    \includegraphics[width=.9\linewidth]{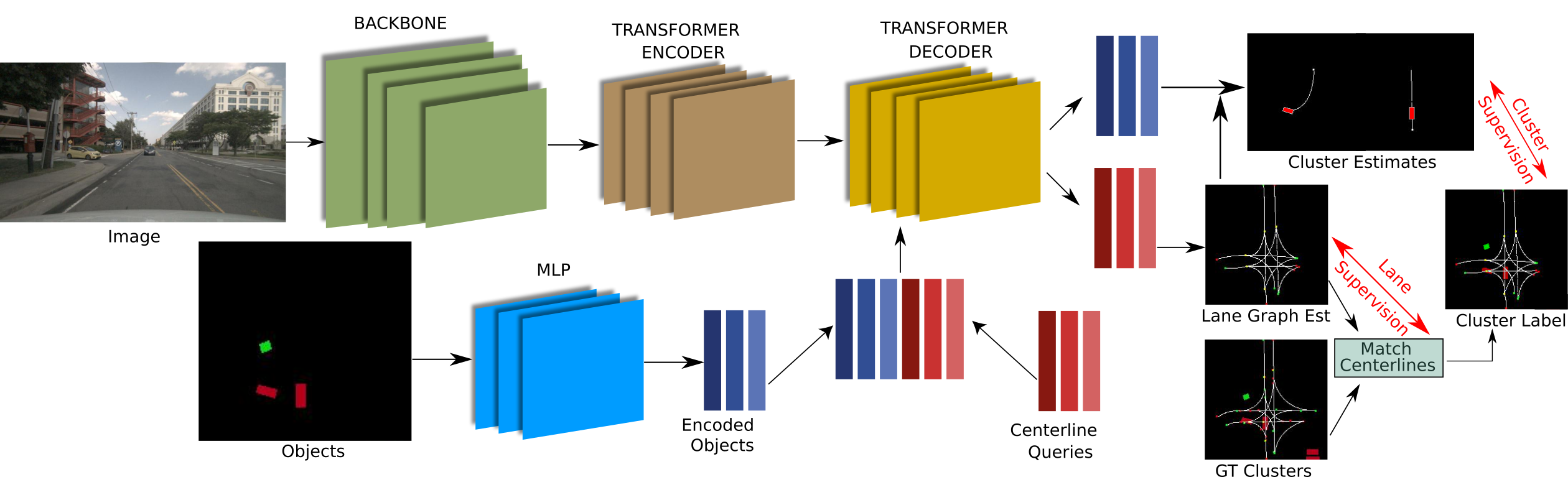}
        \caption{The method accepts an onboard image along with the set of object estimates. The object bounding boxes and semantics are encoded through an MLP. Encoded objects and the learned centerline queries are processed together in a transformer. The centerline queries output the lane graph estimate. Estimated and true centerlines are matched to convert true object cluster memberships into membership labels. }
    \label{fig:method-train}

\end{figure*}

 Here we define an appropriate \textbf{membership function (Z)} that takes the lane graph and objects and outputs a probability distribution. We define the true membership function $\textbf{Z}^*$ as follows: $P(\mathbf{Z}^*|\mathbf{\overline{X}},B,I)$ = $P(\mathbf{Z}^*|\mathbf{\overline{X}},B)$, \ie independent of the image given the lane graph and the objects. Let us define $C_j$ as the 2D center location of the object $B_j$ and $D_{ij}=D(X_i, C_j)$ as the L2 distance between the center $C_j$ and the curve $X_i$. We define the closest curve for object $B_j$ as $M_j = \text{argmin}_{i} D_{ij}$. We also introduce an \textbf{outlier set} $\overline{X}_A$. Then, given the short bounding box side length $W_j$ of the object $B_j$,
\begin{equation}
\medmuskip=-2mu
\scriptsize
P(\mathbf{Z_{ij}^*}|\mathbf{\overline{X}_i},B_j)= 
\begin{cases}
    1,& \text{if} \quad (i\neq A) \& (M_j = i) \& (D_{ij} < W_j)\\
   0,& \text{elif} \quad i\neq A,\\
   1,& \text{otherwise.}
\end{cases}   \label{eq:membership}
\end{equation}
This forms a valid probability function. Essentially, if there is a curve $X_i$ where a given object $B_j$ is closest to and the distance is lower than the short side length of the object $W_j$, the probability function is set to 1 for that entry, \ie $Z_{ij}=1$ and $Z_{kj}=0, k\neq i$. If no such curve exists, then $Z_{Aj}=1$ and $Z_{kj}=0, k\neq A$. This means if the object is not on a centerline, it is assigned to the outlier set $\overline{X}_A$.

Since true membership function $\textbf{Z}^*$ is defined on the true lane graph $\mathbf{\overline{X}}$, it cannot be directly used as a label in loss formulation. As shown in Eq.~\ref{eq:newpost}, the distribution to be used as a label is $P(\overline{Z}|B,\overline{X}, \mathbf{X}, I)$. The cluster membership \textbf{target} distribution is obtained by using the true membership function $\textbf{Z}^*$ on the matching of the true lane graph $\mathbf{\overline{X}}$ and the estimated graph $\mathbf{X}$. To this end, let us define the bipartite matching between centerlines in estimated lane graph $\mathbf{X}$ and the true lane graph $\mathbf{\overline{X}}$ with $H_m$ with $m \in |\mathbf{X}|$, where $H$ takes the index of an estimated centerline and outputs the true centerline it is matched to. We define the estimated equivalent of the tru
e outlier set $\overline{X}_A$ as $X_A$, with $H_A=A$, which means by default, the outlier set of the true lane graph is matched to the outlier set of the estimated lane graph. Then,
\begin{align}
      P(\mathbf{\overline{Z}_{ij}}|\mathbf{\overline{X}},\mathbf{X_i}, B_j)= 
\begin{cases}
    1,& \text{if} \quad \mathbf{Z_{H_{ij}}^*} = 1 \\
   0,& \text{otherwise.}
\end{cases}\,
    \label{eq:targetmembership}
\end{align}
Eq.~\ref{eq:targetmembership} states that the target membership function for an estimated curve $\mathbf{X_i}$ and an object $B_j$ is 1 if object $B_j$ belongs to the true curve that the estimated curve $\mathbf{X_i}$ is matched to. This process is also visualized in Fig.~\ref{fig:objectmatching}. 

\subsection{Architecture}

The method accepts a front facing onboard camera image and a set of 3D object detection estimates and outputs the local BEV lane graph as well as the memberships of the input object estimates to estimated centerlines. To his end, we use a DETR-like \cite{DBLP:conf/eccv/CarionMSUKZ20} architecture as shown in Fig.~\ref{fig:method-train}. The image is processed by a backbone and a transformer encoder. For the positional embeddings on image features, we use the image normalized locations. The input 3D object detection boxes are processed by an MLP to produce per detection feature vectors. The input to the MLP consists of a concatenation of 3D locations of the center and all 8 corners of the object as well as a scalar variable representing the confidence score for the semantic class of the detection. For the ground truth objects, this value is set to 1. All the 3D locations are normalized to [0,1] in the region-of-interest except the height dimension which we do not normalize.

The encoded objects and learnt query vectors are processed jointly in the transformer decoder. The transformer processed centerline queries output i) probability of existence, ii) Bezier control points locations and iii) association features. These outputs are processed in a similar fashion to~\cite{Can_2021_ICCV, can2022topology} to form the lane graph. 

In order to create the target membership distribution explained in Sec.~\ref{sec:membership}, we use Hungarian matching on the control points and probability of existence of the estimated centerlines and the true centerlines. The processed encoded objects output a probability distribution over the centerline estimates. Let us represent the number of centerline estimates, which is the same as the number of centerline queries, with $N_X$ and number of objects with $N_B$. Then Object Cluster Estimates (OCE), as shown in Fig.~\ref{fig:method-train}, is of size $N_B \times (N_X + 1)$ with the additional outlier set. We use softmax on the output of an MLP to ensure $\sum_i \text{OCE}_{ji} = 1, \forall j \in B$, \ie we obtain a valid probability distribution over the centerlines for each object. The OCE is supervised with cross-entropy using target distribution $\overline{Z}$ as a label. We assign a weight of 0.1 on the outlier set since most of the objects are actually not on a centerline. Therefore, the total loss $\mathcal{L} = \mathcal{L_X} + \alpha \mathcal{L_C}$, with $\mathcal{L_X} = \text{L}(X,  \overline{X})$ representing lane graph losses presented in \cite{Can_2021_ICCV} and $\mathcal{L_C}$ is the clustering loss with trade-off hyperparameter $\alpha$. If there are no objects in the sample, then object clustering loss is simply set to zero for that sample. Since the input query vectors are concatenations of the encoded objects and the learnt centerline query vectors, if there is no object detection for a sample, the input to the transformer decoder is only the learnt centerline queries.

Given the architecture and the theoretical explanation of the method, implementation is presented in Alg.~\ref{al:findManhattanFrame}. The algorithm outlines how to obtain the label membership function, \ie the centerline each object is assigned given the image, objects, the model and an outlier set. True membership is obtained through the implementation of Eq.~\ref{eq:membership} which uses the object detection box center and short side length.
The estimated and true centerlines are matched through the Hungarian algorithm. Note that $H$ represents the function that takes the estimated centerline index and outputs the true centerline it is matched to as explained in Eq.~\ref{eq:targetmembership}.  $H'$ is the related function where it takes the index of the true centerline and outputs the estimated centerline it is matched to. In practice, $H$ and $H'$ are simply different indexing of the Hungarian matching output. The true membership is converted to the form of estimated centerlines through Hungarian matching. The label memberships are used to supervise the model. In Step 9 of Alg.~\ref{al:findManhattanFrame} we minimize the loss with gradient descent. Comparing this with Eq.~\ref{eq:probmax}, it can be seen that maximizing the total term, which is composed of log likelihoods, is the same as minimizing the corresponding losses in Step 9 of Alg.~\ref{al:findManhattanFrame}. The posterior term in Eq.~\ref{eq:tempmax} translates to the negative of the lane graph loss while the clustering term is the negative of the cross entropy loss $\mathcal{L_C}$. 

\begin{algorithm}
{
\small
\caption{$\theta_{n+1}$ = OptimizeParams($I,B,F_{\theta_n(.)}, \overline{X}, A, \alpha$)}
\label{al:findManhattanFrame}
\begin{algorithmic}
 \STATE 1. Infer $P(Z), X = F_{\theta}(I,B)$.
 \STATE 2. ${H(m) = \text{Hungarian}(X, \overline{X})}, {H'(n) = \text{Hungarian}(\overline{X},X)}$.
 \STATE 3. Compute pair-wise distances $D(i,j) = ||\overline{X}_i - C_j||_2, \forall i,j$. 
 \STATE 4. Find the closest pairs such that $M_j = \text{argmin}_i D(i,j)$.
 \STATE 5. If $D(C_j,j) <= W_j$, set $Z^*_j = M_j, \overline{Z}_j = H'(j)$,\\
 -- Otherwise, set $Z^*_j = A, \overline{Z}_j = A$.    
 \STATE 6. Computer clustering loss $\mathcal{L_C} = CE(P(Z),  \overline{Z})$
 \STATE 7. Computer lane graph loss $\mathcal{L_X} = \text{L}(X,  \overline{X})$.
 \STATE 8. Set loss total loss to $\mathcal{L} = \mathcal{L_X} + \alpha \mathcal{L_C}$.
 \STATE 9. Update params $\theta_{n+1} = \theta_{n} - \partial\mathcal{L}/\partial \theta$. \\
 -- Note: minimization loss equates to maximizing $\mathbf{TT}$ in~\eqref{eq:probmax}.
 \STATE 10. Return $\theta_{n+1} $ .
\end{algorithmic}
}
\end{algorithm}

\noindent \textbf{Object refinement (RE).} We also experiment with an optional additional MLP that takes the transformer processed object vectors and outputs the BEV locations for the object centers. The module is supervised by the GT object center locations with an L1 loss. The output object bounding boxes are obtained by only replacing object center locations of the input bounding boxes with the estimated center locations and keeping the orientation, lengths and class identical.

\section{Experiments}

We use the NuScenes~\cite{nuscenes2019} and Argoverse~\cite{DBLP:conf/cvpr/ChangLSSBHW0LRH19} datasets. The object detection task use the canonical train/val split while in BEV scene understanding literature a different split (BEV Split) is commonly used \cite{DBLP:conf/cvpr/RoddickC20, Can_2021_ICCV, can2022topology, xu2022centerlinedet}. In order to validate the applicability of the proposed method, we use the estimations of some top scoring 3D object detection methods in the used datasets at the time of the paper submission. Therefore, we report our results in canonical split by training the competitor methods in the canonical split. We only use CBGS\cite{DBLP:journals/corr/abs-1908-09492} for Argoverse dataset due to a lack of methods with an official leaderboard entry. For NuScenes, we use DeepInteraction\cite{DBLP:journals/corr/abs-2208-11112}) and BevFusion\cite{DBLP:journals/corr/abs-2205-13542}, both of which are among the top scoring methods in official leaderboard. We also experiment with using the GT objects. In order to show the effectiveness of the proposed object clustering loss, we experiment training all methods with and without this loss, where we put the suffix \textbf{OC} to indicate methods that use the clustering loss. Moreover, we investigate the 3D object detection performance of our method by training it to estimate the 3D center of the given 3D object detections. We refer to models trained to refine object locations with \textbf{RE}. We do not apply refinement on the models that take GT objects as input. For the refinement methods we also experiment with not training them to extract lane graph, in which case the model is only trained to refine the locations of objects. We call the lane graph training with \textbf{LG}. For the metrics we use the same set of measures as \cite{Can_2021_ICCV} which cover the centerline accuracy as well as the connectivity.

\noindent\textbf{Implementation.}
The BEV target area that our method outputs the lane graph for is from -25 to 25m in x-direction and 1 to 50m in z direction. The backbone network is Deeplab v3+ \cite{DBLP:conf/eccv/ChenZPSA18} pretrained on Cityscapes dataset \cite{Cordts2016Cityscapes}. Our implementation is in Pytorch and runs with 10FPS. 

\noindent\textbf{Baselines.}
We compare the performance of our method against the state-of-the-art~\textbf{STSU}~\cite{Can_2021_ICCV} and \textbf{TPLR}~\cite{can2022topology}.

\section{Results}

We present our results against state-of-the-art methods in NuScenes and Argoverse datasets and provide ablation studies on the NuScenes dataset.

\subsection{Quantitative Results}











The results of our method in NuScenes are given in Tab.~\ref{tab:detailed}. The first observation is that all models of our method outperform the SOTA. Using the proposed architecture without object clustering already performs significantly better than existing methods. Introducing the object clustering loss provides a performance boost with both object detection methods as well as GT object detections. Furthermore, training the network to refine the center locations of object detections further improves the results. However, the refinement results of the method is lower than the object detection methods' results with BEVFusion's mAP being 68.5 and DeepInteraction's 69.8. Comparing the methods with and without \textbf{LG}, it can be seen that lane graph supervision is helpful in object center refinement performance. However, our work focuses on improving lane graph accuracy through the utilization of object detections rather than vice versa. The refinement loss helps more when object clustering is also applied. Comparing GT results with object detection methods, it can be seen that GT performs the best in both \textbf{LG} and \textbf{LG+OC} methods. However, the addition of \textbf{RE} pushes the performance of the other methods above the GT trained model. The final observation is that the results of a particular component setting are very similar to each other across the object detection methods and GT. This indicates that the proposed method can work with different object detection methods. 

The results for Argoverse are given in Tab.~\ref{tab:argo}. On the Argoverse dataset, our method significantly outperforms the state-of-the-art methods. Furthermore, in both GT objects and CBGS estimates cases, the proposed object clustering formulation provides substantial improvements in all metrics. These observations confirm the results in NuScenes.

\subsection{Ablations}

Since our method uses object detections rather than the intermediate representations of the object detection methods, the models should be robust against the use of different object detection methods in training and test time. To validate this, we run experiments on the Nuscenes dataset. As can be seen in Tab.~\ref{tab:domain}, we test all models using the object detections of the other methods. The general trend is that the use of GT objects at test time provides a small improvement in the results regardless of what method was used in training. The results being stable across changes in the test time object detection methods validates that the proposed framework provides a robust lane graph extraction model.












\begin{table}[h]
\begin{center}
\footnotesize{
\tabcolsep=0.2cm
\begin{tabular}{ |c|c|c|c|c|c|c|c| }
\hline
 &  \multicolumn{3}{|c|}{Components} & \multicolumn{4}{|c|}{NuScenes} \\
\hline
\textbf{Method} & LG & OC & RE &  M-F &  Detect & C-F & mAP  \\
\hline
\hline
STSU & \ding{51} & & &  61.7 & 64.8  & 52.8 & - \\

TPLR & \ding{51} & & &  62.6 & 65.8  & 53.5 & - \\
\hline
DeepInter & \ding{51} & & & 63.2 & 69.6 & 55.3 & -\\

DeepInter&  &  & \ding{51} & - & - & - & 47.8 \\

DeepInter& \ding{51} & \ding{51} & & 64.2 & 69.6 & 56.2 & -\\

DeepInter& \ding{51} &  & \ding{51} & 63.2& 69.9 & 56.2 & 58.6\\

DeepInter& \ding{51} & \ding{51} & \ding{51} & \textbf{64.8} & \textbf{70.6}  & \textbf{57.4} & \textbf{58.9}\\

\hline

BEVFuse & \ding{51} & & & 63.4 & 69.2 & 56.2 & -\\

BEVFuse&  &  & \ding{51} & - & - & - & 51.6 \\

BEVFuse& \ding{51} & \ding{51} & & 63.9 & 69.3 & \textbf{56.5} & -\\

BEVFuse& \ding{51} &  & \ding{51} & 63.7 & 69.3 & 55.2 & 56.5\\

BEVFuse& \ding{51} & \ding{51} & \ding{51} & \textbf{64.9} & \textbf{69.5} & 56.3 & \textbf{56.8}\\

\hline

GT & \ding{51} & & & 63.8 & 70.6 & 56.7 & -\\

GT& \ding{51} & \ding{51} & & \textbf{64.2} & \textbf{70.6} & \textbf{57.2} & -\\

\hline
\end{tabular}
}

\end{center}
\caption{Detailed results on NuScenes dataset. Performance of models with different components and object detection method outputs are given. }
\label{tab:detailed}
\end{table}

\begin{table}[h]
\begin{center}
\footnotesize{
\tabcolsep=0.25cm
\begin{tabular}{ |c|c|c|c| }
\hline
&  \multicolumn{3}{|c|}{Argoverse} \\
\hline
\textbf{Method} & M-F &  Detect & C-F  \\
\hline
\hline


STSU \cite{Can_2021_ICCV}& 64.8 & 63.0 & 61.6\\

TPLR \cite{can2022topology}& 65.8 & 69.2 & 63.0\\
\hline
Ours(CBGS\cite{DBLP:journals/corr/abs-1908-09492})  &67.6 & 70.8 & 63.8\\

Ours-OC(CBGS\cite{DBLP:journals/corr/abs-1908-09492})  &\textbf{68.9} & \textbf{71.7} & \textbf{64.9}\\
\hline
Ours(GT)  & 67.8 & 71.0 & 63.6\\

Ours-OC(GT)  &\textbf{69.2} & \textbf{71.0} & \textbf{63.6}\\

\hline
\end{tabular}
}

\end{center}
\caption{Argoverse results.}
\label{tab:argo}
\end{table}

\begin{figure*}
    \centering
    \includegraphics[width=\linewidth]{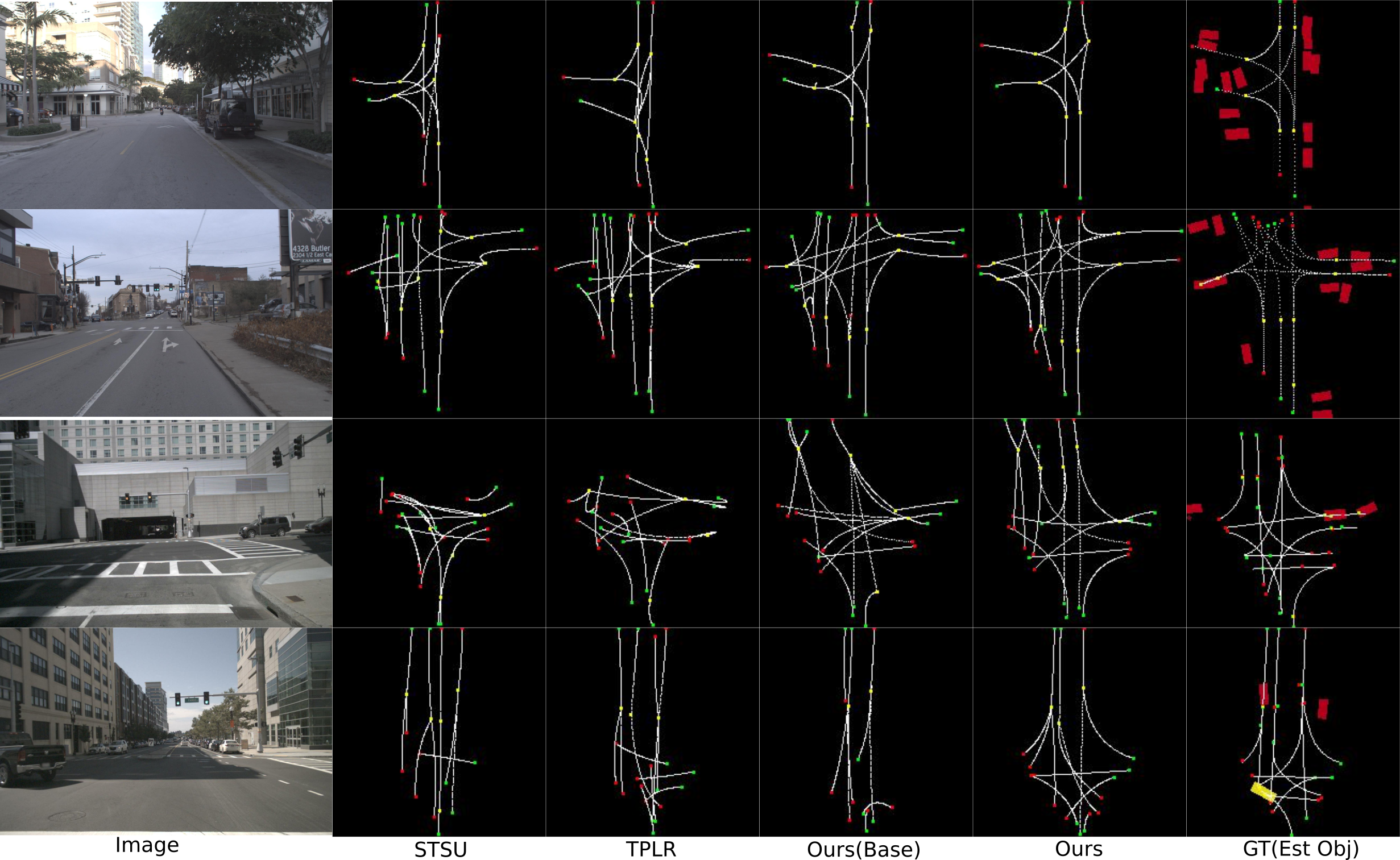}
        \caption{Visual results on Argoverse (top 2 rows) and NuScenes (bottom 2 rows). GT lane graphs are presented with detected objects from CBGS\cite{DBLP:journals/corr/abs-1908-09492} for Argoverse and BEVFusion \cite{DBLP:journals/corr/abs-2205-13542} for NuScenes.}
    \label{fig:visres}
\end{figure*}

\begin{table}[h]
\begin{center}
\footnotesize{
\tabcolsep=0.25cm
\begin{tabular}{ |c|c|c|c| }
\hline
& \multicolumn{3}{|c|}{Tested with (M-F results)} \\
\hline
\textbf{Trained} & BevFusion &  DeepInter & GT   \\
\hline
\hline

BevFusion\cite{DBLP:journals/corr/abs-2205-13542} & 64.9 & 64.8 & 64.9 \\

DeepInteraction\cite{DBLP:journals/corr/abs-2208-11112} & 64.7 & 64.8 & 64.8 \\

GT & 64.1 & 64.1 & 64.2 \\

\hline
\end{tabular}
}

\end{center}
\caption{Study of models' performance when trained and tested with different object detection methods on Nuscenes (M-F score). }
\label{tab:domain}
\end{table}

The results for the object membership estimation accuracy is given in Tab.~\ref{tab:cluster}. All models excel at estimating membership with the object detections they are trained with. The models trained with object detection methods perform worse when given the true objects while the difference in accuracy for GT trained model is smaller. Generally, all models demonstrate a high level of accuracy in object membership estimation.

\begin{figure*}
    \centering
    \includegraphics[width=\linewidth]{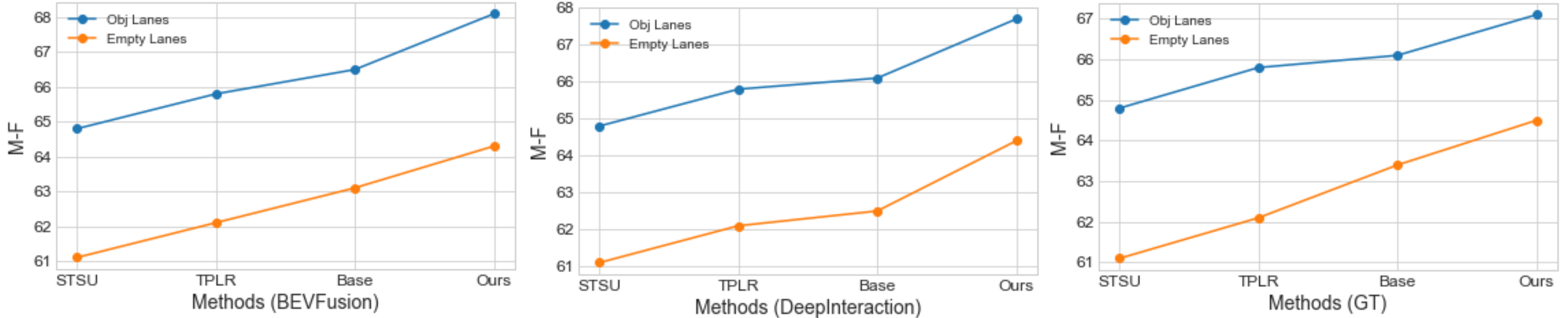}
        \caption{M-F score of the methods on the centerlines with objects assigned to them (Obj Lanes) and centerlines without objects (Empty Lanes). Proposed formulation provides large boost in Obj Lanes.}
    \label{fig:graphs}
\end{figure*}

We further dissect the performance of the methods into the centerlines with and without objects as given in Fig.~\ref{fig:graphs}. The results show that in all object detection methods, the proposed object clustering formulation provides a large boost in performance in centerlines with objects. Moreover, it can be seen that the performance also increases for centerlines without objects.

\begin{table}[h]
\begin{center}
\footnotesize{
\tabcolsep=0.25cm
\begin{tabular}{ |c|c|c|c| }
\hline
& \multicolumn{3}{|c|}{Tested with (Object Clustering Accuracy)} \\
\hline
\textbf{Trained} & BevFusion &  DeepInter & GT   \\
\hline
\hline

BevFusion\cite{DBLP:journals/corr/abs-2205-13542} & 90.3 & 90.0 & 83.9 \\

DeepInteraction\cite{DBLP:journals/corr/abs-2208-11112} & 90.2 & 90.4 & 84.2 \\

GT & 89.5 & 89.4 & 91.8 \\

\hline
\end{tabular}
}

\end{center}
\caption{Study of models' accuracy in assigning the objects to centerlines when trained and tested with different object detection methods on Nuscenes dataset. }
\label{tab:cluster}
\end{table}

\subsection{Visual Results}

The visual results for STSU \cite{Can_2021_ICCV}, TPLR \cite{can2022topology}, our method without object clustering (Ours(Base)) and Our method with object clustering (Ours) are given in Fig.~\ref{fig:visres}. The top two rows show the results for the Argoverse dataset where the object detections are from CBGS \cite{DBLP:journals/corr/abs-1908-09492}. Our methods perform significantly better than the competitors. It can also be seen that the locations of the objects correlate with the improvements of our method compared to the competitors showcasing the effectiveness of the proposed framework. 

In the bottom two rows of Fig.~\ref{fig:visres}, we present the visual results for NuScenes. The object detections are from BEVFusion \cite{DBLP:journals/corr/abs-2205-13542}. In both samples from NuScenes, both of our methods provide better estimates than the competitors. Comparing Ours(Base) with Ours, the improvement from the proposed object clustering is also clearly visible.

\section{Conclusion}

In this work, we address the task of online lane graph extraction from an onboard camera image. We propose to utilize the onboard object detection algorithms to improve the accuracy in lane graph estimation. To this end, a new architecture that takes an input image and a set of object detections is proposed. The proposed architecture is trained with a novel formulation that couples the object sand the centerlines of the lane graph. In order to achieve this coupling, we derive a clustering term that assigns the objects to the centerlines. This clustering term is used as a loss to supervise the model. Our extensive experiments in NuScenes and Argoverse datasets show clear improvement over the state-of-the-art. Moreover, the ablation studies demonstrate that our approach can successfully assign objects to the centerlines. Our method is robust against use of different object detection methods in train and test times, allowing a deployed model of our method to work with different object detection pipelines without need for retraining. 

\noindent \textbf{Limitations.} The cases with objects not following the centerline but being close to the centerlines such as lane changing or road side parking are challenging for the method. These cases require further research in the future works.

\noindent\paragraph{Acknowledgements.} The authors gratefully acknowledge support from Toyota Motors Europe (TME), and the Ministry of Education and Science of Bulgaria (for INSAIT, part of the Bulgarian National Roadmap for Research Infrastructure).

{\small
\bibliographystyle{ieee_fullname}
\bibliography{egbib}
}

\newpage

\section{Supplemantary-Formulation}

In this section, we provide the details regarding our formulation and the Expectation-Maximization paradigm. Let us start by remembering the initial probability we want to maximize in Eq 2:
\begin{align}
    P(\mathbf{X}| B, I) \propto P(B|\mathbf{X}, I)P(\mathbf{X}|I)\,.
    \label{eq:eq2}
\end{align}. 

In the end of our derivation we arrived in the following \textbf{total term}:
{\medmuskip=-0mu
\small
\begin{align}
    \mathbf{TT} = \overbrace{\text{log}P(\mathbf{X}|B,I)}^\text{Posterior Term} + \overbrace{\sum_{z}P(\mathbf{Z}|B,X_n,I) \text{log}P(\mathbf{Z}| B, \mathbf{X},I)}^\text{Clustering Term}.
    \label{eq:tempmax}
\end{align}
}

To better understand the clustering term, let us create a new term by adding $P(B|\textbf{X},I)$ to the clustering term,
{\medmuskip=-0mu
\small
\begin{align}
    \sum_{z}P(\mathbf{Z}|B,X_n,I) \text{log}P(\mathbf{Z},B| \mathbf{X},I) = \nonumber \\
    \text{log}P(B| \mathbf{X}, I) + \sum_{z}P(\mathbf{Z}|B,X_n,I) \text{log}P(\mathbf{Z}| B, \mathbf{X},I).
    \label{eq:emtemp}
\end{align}
}

We recognize the term in Eq \ref{eq:emtemp} as the expected value of the log likelihood function in expectation-maximization paradigm. Specifically, the set of data points is $B$, the latent variable is $\mathbf{Z}$ and the parameters to maximize is given by $\mathbf{X}$ with the parameter values in current iteration is $X_n$. Then, let us name the log likelihood term $ \sum_{z}P(\mathbf{Z}|B,X_n,I) \text{log}P(\mathbf{Z},B| \mathbf{X},I) $ as \textbf{EM} term. Then, \textbf{total term} becomes

{\medmuskip=-0mu

\begin{align}
\text{log}P(\mathbf{X}|B,I) -  \text{log}P(B| \mathbf{X}, I) + \mathbf{EM}.
    \label{eq:comb}
\end{align}
}

If we rearrange the terms, we arrive at the following formula:

{\medmuskip=-0mu

\begin{align}
\text{log}P(\mathbf{X}|I) -  \text{log}P(B| I) + \mathbf{EM}.
    \label{eq:comb2}
\end{align}
}

Since $P(B| I)$ is given and does not depend on $\mathbf{X}$, we can remove it from the formula. We see that through our formulation, we actually maximize \textbf{EM} term and $\text{log}P(\mathbf{X}|I)$. Comparing with \ref{eq:eq2}, we see that we directly maximize $P(\mathbf{X}|I)$ and apply expectation maximization on $P(B|\mathbf{X}, I)$. This is evident from the fact that \textbf{EM} term is exactly the log likelihood for data points $B$ and parameters $\mathbf{X}$. 

In practice we feed the object information to the network, thus we actually maximize the \textbf{target distribution TD}:
{\medmuskip=-0mu
\begin{align}
\mathbf{TD} = \text{log}P(\mathbf{X}|I, B) + \mathbf{EM}.
    \label{eq:fin}
\end{align}
}

\textbf{TD} has the traditional posterior term as well as the \textbf{EM} term. As shown in our experiments, EM term provides additional supervision through explicit modelling of the relationship between the objects and the centerlines.

\section{Additional results}

\begin{figure*}
    \centering
    \includegraphics[width=.9\linewidth]{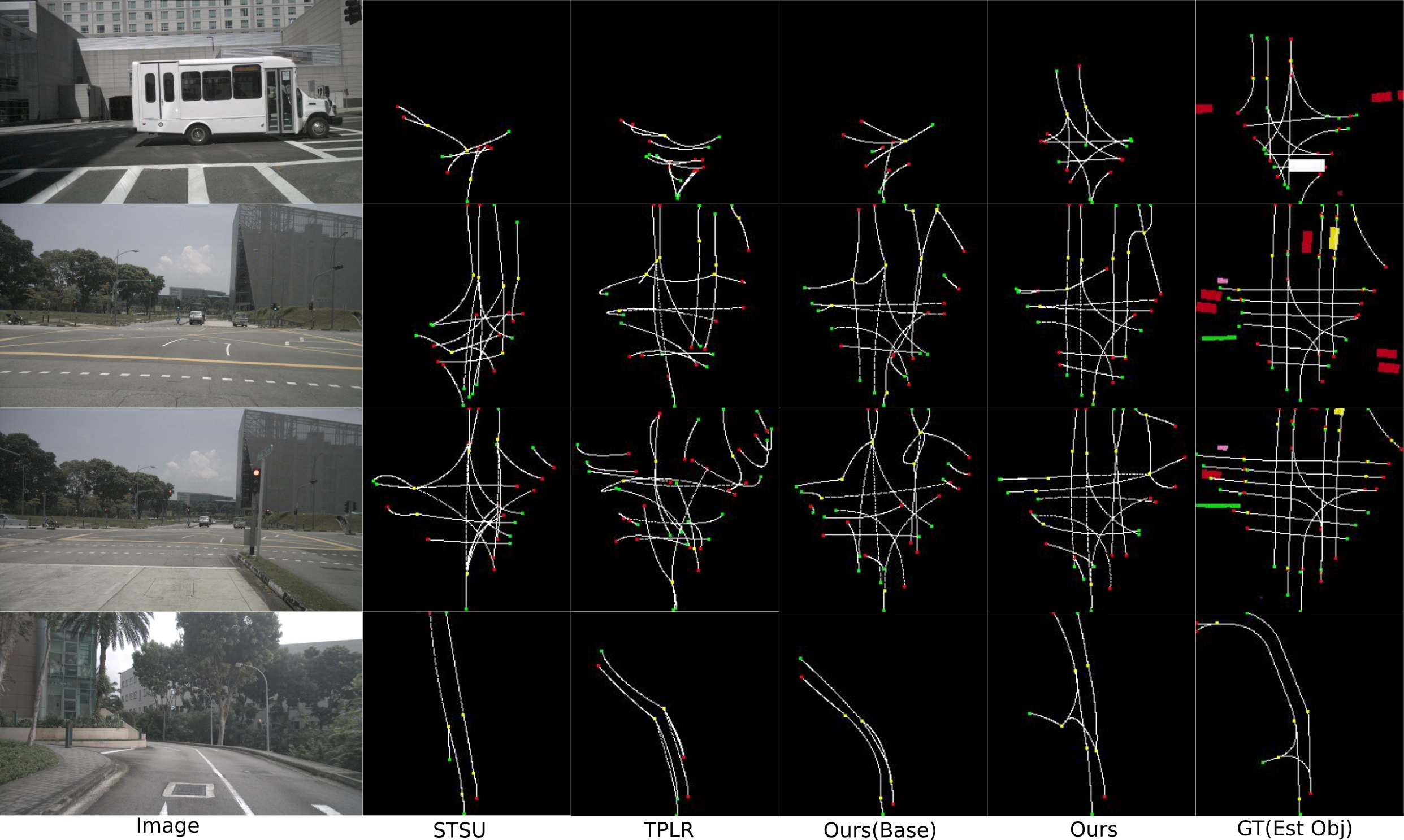}
        \caption{Visual results on NuScenes dataset. Top 2 rows are from DeepInteraction method while the bottom 2 rows are from the method with GT objects. It can be seen that even if there is no object in the scene, the method trained with the proposed formulation performs better.}
    \label{fig:graphs2}
\end{figure*}

In this section, we provide some additional visual results on NuScenes dataset with DeepInteraction and GT methods in Fig \ref{fig:graphs2}. In both settings, the base version of our method outperforms the SOTA while the model trained with the proposed object clustering formulation provides even better results. Another important observation is about the last row of the gifure where there is no objects in the scene. We see that the model with OC (Ours) performs much better than all other methods even in the absence of objects. This shows that the proposed framework leads to a model with better understanding of the road topology.

\end{document}